# UNDITHERING USING LINEAR FILTERING AND NON-LINEAR DIFFUSION TECHNIQUES

**V. Asha**

Department of Master of Computer Applications, New Horizon College of Engineering,

Bangalore – 560 103, India

Email: v_asha@live.com

**ABSTRACT**

*Data compression is a method of improving the efficiency of transmission and storage of images. Dithering, as a method of data compression, can be used to convert an 8-bit gray level image into a 1-bit / binary image. Undithering is the process of reconstruction of gray image from binary image obtained from dithering of gray image. In the present paper, I propose a method of undithering using linear filtering followed by anisotropic diffusion which brings the advantage of smoothing and edge enhancement. First-order statistical parameters, second-order statistical parameters, mean-squared error (MSE) between reconstructed image and the original image before dithering, and peak signal to noise ratio (PSNR) are evaluated at each step of diffusion. Results of the experiments show that the reconstructed image is not as sharp as the image before dithering but a large number of gray values are reproduced with reference to those of the original image prior to dithering.*

Keywords: Dithering; Average filter; Anisotropic diffusion; First-order statistics; Second-order statistics

## 1. INTRODUCTION

Methods of compressing the data prior to storage and/or transmission are of significant and commercial interest. Image compression addresses the problem of reducing the amount of data required to represent a given quantity of information and is a key technology in various multimedia services, document and medical imaging, and military and space application. The compression process is applied prior to storage or transmission of the image. Later the compressed image is decompressed to reconstruct an image which is the approximation of the original image. The reconstruction may be lossy or lossless depending on the compression method used. In a lossless data compression system, the recovered image is identical to the original image. The goal in a lossy compression system is to reconstruct an image which resembles the original image as close as possible at a lowest possible bit rate. Compression ratios are much higher for lossy compression than for lossless compression. Dithering is a technique to display more number of gray levels on a device with black and pixels only [1]. Dithering has been widely used in areas in medical imaging [2] and printing industries [3], [4], [5]. Dithering of an eight-bit image to a one-bit (binary) image compresses the image at a compression ratio of 1:8 resulting in a lossy compression [6]. The two most common methods are ordered dither and Floyd-Steinberg dither. Ordered dither uses a cleverly chosen set of black-and-white patterns to represent different gray values using a thresholding scheme to replace each gray pixel with a black or white pixel [7]. Floyd-Steinberg dither is an error diffusion algorithm that processes the pixels of each line in an image from left to right and top to bottom [8]. Each pixel is examined and rounded off to either black or white by compensating the error to the neighboring pixels such that the information is not lost. This method is considered to be better than ordered dither as it suits well for representing fine lines. The reason why dithered images appear as continuous gray shade images to human vision system is that the human eyes automatically blur the dots into gray shades [9]. While other compression methods are considered, there are two stages, namely, compression and decompression or reconstruction. In case of dithering, only one stage is achieved. Undithering aims at conversion of dithered image back to original image as close as possible. Using undithering, it is possible to convert a 1-bit image to an 8 bit-image [9] or an 8-bit image to a 24-bit image [10]. In Stenger's method of undithering for reconstruction of an 8-bit image from a 1-bit image, the first step is to blur the image by replacing each pixel's value with average of pixels in a small window around the centre pixel with the help of linear filter. Since the linear filtered image appears mottled due to less number of gray levels, the next step involves non-linear smoothing using Lee's local statistics [11] followed by sharpening. Thus, an undithering algorithm should involve computerized smoothing or blurring, smoothing and edge enhancement. In the present method, I propose a method of undithering using linear filtering and anisotropic diffusion that combines the advantage of smoothing and edge enhancement. Linear filtering prior to anisotropic diffusion, as a pre-processing, involves smoothing and diffusion of information within the window of selected size. Anisotropic diffusion that involves adaptive smoothing is then applied on the linear-filtered image at controlled rate and several features are evaluated at each iteration step.

## 2. BRIEF REVIEW ON DIFFUSION

The basic idea of diffusion in image processing arose from the heat diffusion equation. The rate at which temperature at any point (x,y) in a two-dimensional field changes with time t can be given as,

$$\frac{\partial T(x,y,t)}{\partial t} = \nabla \cdot \left(\alpha \nabla T(x,y)\right) \qquad (1)$$

where α is the thermal diffusivity, $\nabla$ is the gradient operator and $\nabla \cdot$ is the divergence operator. Anisotropic diffusion in image processing is a discretization of the family of continuous partial differential equations that include both the physical processes of diffusion and the Laplacian [12]. The same equation is applicable to image function f(x,y) as,

$$\frac{\partial f(x,y)}{\partial t} = \nabla \cdot \left(c \nabla f(x,y)\right) \qquad (2)$$

where c is the image diffusivity.

If c is constant, independent of space, it leads to a linear diffusion equation with homogeneous diffusivity, in which case, all pixels in the image, including sharp edges and corners, will be blurred at a uniform rate.

A simple modification here would be choosing image diffusivity as a function of image gradient itself. In such case, the diffusion becomes non-linear diffusion and the gradient function becomes an “edge-stopping” function. The diffusion near the edges and other areas with rapidly varying gray levels has to be minimal and that away from those areas has to be maximal. A qualitative description of this is shown in Fig. 1 [12].

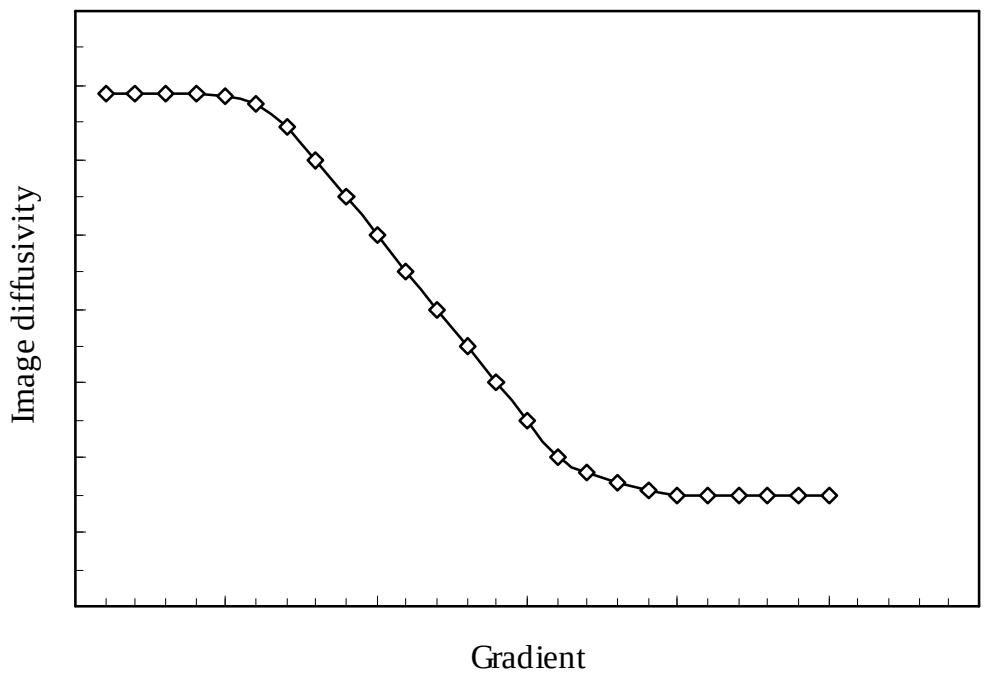

Fig. 1. Qualitative representation of non-linearity of image diffusivity

Various applications of diffusion include de-noising [13], [14], [15], [16], segmentation [17], [18], inpainting [19] and enhancing image resolution (zooming) [20]. A critical issue in the diffusion process is the good choice of diffusivity function. Several authors have proposed diffusivity function in various forms. In the present work, I intend to choose diffusion function which is of the form as below [21]:

$$c(x, y) = \frac{1}{\left|\nabla f(x, y)\right|^{p}} \quad (3)$$

Diffusivity functions of such type will lead to numerical problems when the gradient gets close to zero. This problem can be avoided by adding a small positive constant ε to the denominator. For p = 1, the diffusivity function becomes total variation (TV) flow [22], [23], a diffusion filter that is equivalent to TV regularization [24], [25]. This function seems to suit very well, since it removes oscillations, resulting in piecewise constant results.

## 3. EVALUATION OF FEATURES

First order statistics, second order statistics, MSE and PSNR are considered to be the features for evaluation at each diffusion step. The first order statistics is based on probability of occurrence of a gray level $r_k$ in an image given by

$$p(r_k) = \frac{n_k}{n}, \ k = 0, 1, 2, \dots, L-1 \quad (4)$$

where, n is the total number of pixels, $n_k$ is the number of pixels that have gray level $r_k$, and L is the total number of gray levels in the image. The properties calculated are mean ($\bar{r}$), variance ($\mu_2$), skewness ($\mu_3$), kurtosis ($\mu_4$), energy ($E_1$), and entropy ($S_1$) and are given as below:

$$\bar{r} = \sum_{k=0}^{L-1} r_k p(r_k) \tag{5}$$

$$\mu_2 = \sum_{k=0}^{L-1} (\bar{r} - r_k)^2 p(r_k) \tag{6}$$

$$\mu_3 = \sum_{k=0}^{L-1} (\bar{r} - r_k)^3 p(r_k) \tag{7}$$

$$\mu_4 = \sum_{k=0}^{L-1} (\bar{r} - r_k)^4 p(r_k) \tag{8}$$

$$E_1 = \sum_{k=0}^{L-1} [p(r_k)]^2 \tag{9}$$

$$S_1 = -\sum_{k=0}^{L-1} p(r_k) \log_2 [p(r_k)] \tag{10}$$

In order to analyze the reconstructed image based on second order statistics, co-occurrence matrices are constructed based on second order probability $p_{\theta,d}(a,b)$ defined as the probability of occurrence of a gray level ‘a’ with another gray level ‘b’ separated by a distance ‘d’ in the direction ‘θ’ in an image. The properties calculated are energy ($E_2$), entropy ($S_2$), contrast ($C_t$), homogeneity (H), and correlation ($C_n$) and are given below:

$$E_2 = \sum_a \sum_b [p_{\theta,d}(a,b)]^2 \tag{11}$$

$$S_2 = -\sum_a \sum_b p_{\theta,d}(a,b) \log_2 p_{\theta,d}(a,b) \tag{12}$$

$$C_t = \sum_a \sum_b (a-b)^2 p_{\theta,d}(a,b) \tag{13}$$

$$H = \sum_a \sum_b \frac{p_{\theta,d}(a,b)}{1+|a-b|} \tag{14}$$

$$C_n = \frac{\sum_a \sum_b (a-\mu_x)(b-\mu_y)p_{\theta,d}(a,b)}{\sigma_x \sigma_y} \tag{15}$$

where, $\mu_x = \sum_a p_{\theta,d}(a,b)$ (16)

$$\mu_y = \sum_b p_{\theta,d}(a,b) \tag{17}$$

$$\sigma_x = \sum_a (a-\mu_x)^2 \sum_b p_{\theta,d}(a,b) \tag{18}$$

$$\sigma_y = \sum_b (b-\mu_y)^2 \sum_a p_{\theta,d}(a,b) \tag{19}$$

MSE and PSNR are the most common measures of picture quality in image compression systems, though these are not adequate as perceptually meaningful measures [26], [27]. At every iteration step of diffusion process, n, the MSE and PSNR are also calculated as below:

$$MSE = \frac{1}{MN} \sum_{i=1}^{M} \sum_{j=1}^{N} [f_o(i,j) - f_n(i,j)]^2 \tag{20}$$

$$PSNR = 20 \times \log_{10}(255/\sqrt{MSE}) \tag{21}$$

where $f_o$ indicates the original image before dithering and $f_n$ is the reconstructed image at any iteration step n. In general, lower the value of MSE, better is the effectiveness of reconstruction. However, this measure does not necessarily imply that an image with a lower MSE is always visually pleasing.

## 4. EXPERIMENTS AND RESULTS

In order to study the reconstruction process, Floyd-Steinberg dither was applied on two gray images (Pepper and Baboon). The dithered images were twice filtered using average filter of size 3 × 3. The linear filtered images were subjected to anisotropic diffusion with p = 1, ε = 0.001 and Δt = 0.1 for 200 iterations. At each step of diffusion process, all measures described in the previous section were calculated. As far as the co-occurrence matrix properties are concerned, there is hardly any difference among direc-tions 45°, 90°, 135° and 180° at unit pixel displacement. Here second order statistical properties were calculated for θ = 0° and d = 1. Fig. 2 (a) shows the pepper test image. A small portion of this test image is enlarged and the detail is shown in Fig. 2 (b). First order and second order properties and MSE / PSNR for the reconstructed image at every iteration are shown in Fig. 3 and Fig. 4 along with the properties of the original image before dithering. First order properties like mean, variance, skewness and kurtosis slowly vary with the iteration. Beyond certain step, first order energy and entropy and second order energy, contrast and correlation do not vary much, but homogeneity keeps increasing at a faster rate. From the plot of MSE / PSNR, MSE is found to be minimum at a step of 46. However, from aesthetic point of view, an iteration step of 120 yields better visual appearance of the reconstructed image. The reconstructed image is shown against the original image in Fig. 5. The reconstructed image is slightly blurred with reference to original image and it is clear from the details of the windows shown in Fig. 6. From the plot of histograms as in Fig. 7, it is seen that the distribution of gray levels of the reconstructed image is similar to that of the original image before dithering. Typical gray profiles (Fig. 8) show that the gray level variation is also similar for both reconstructed image and original image.

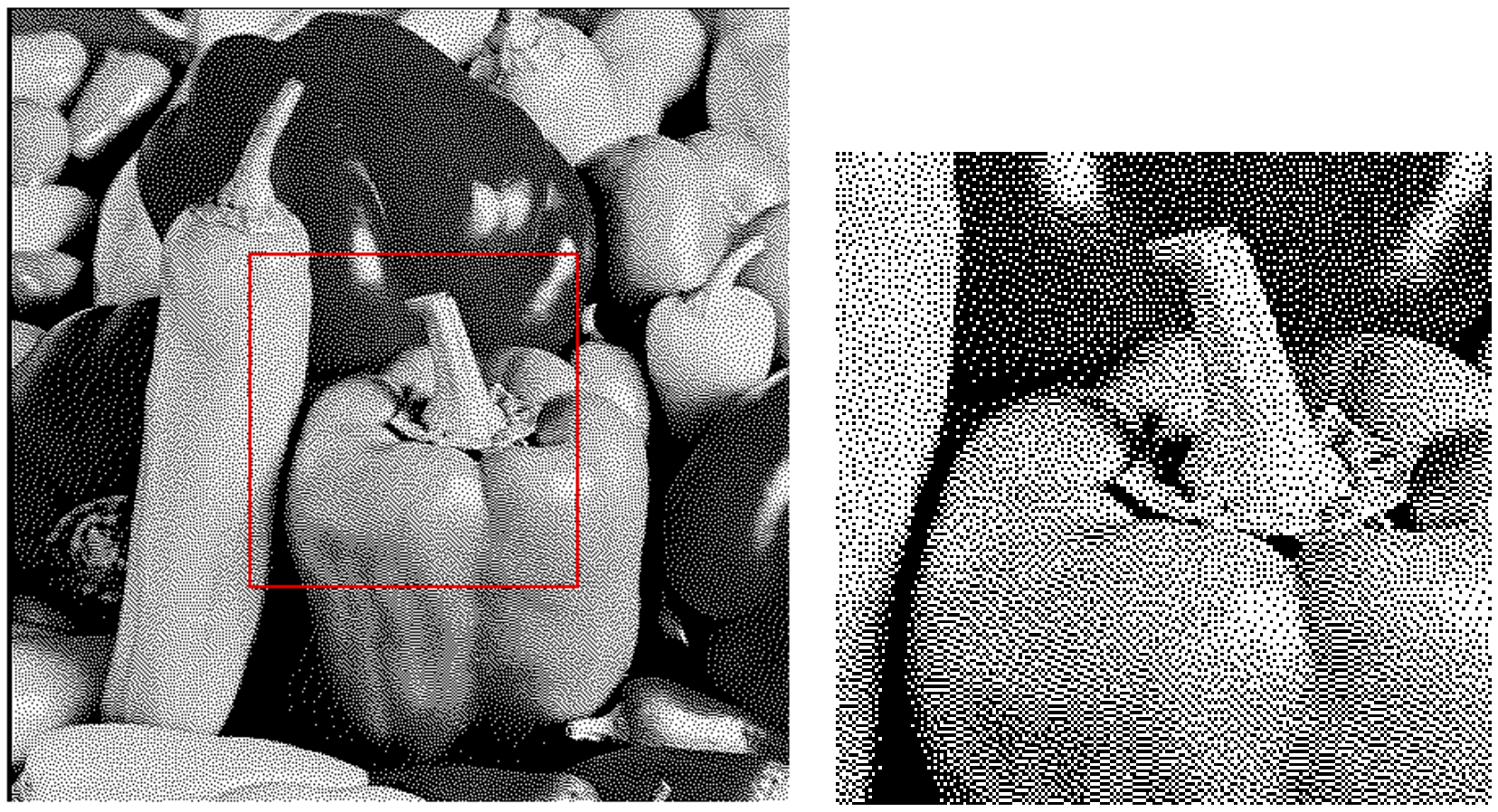

(a) (b)

Fig. 2. Pepper image in dithered form (a) Full view (b) Detail of the window

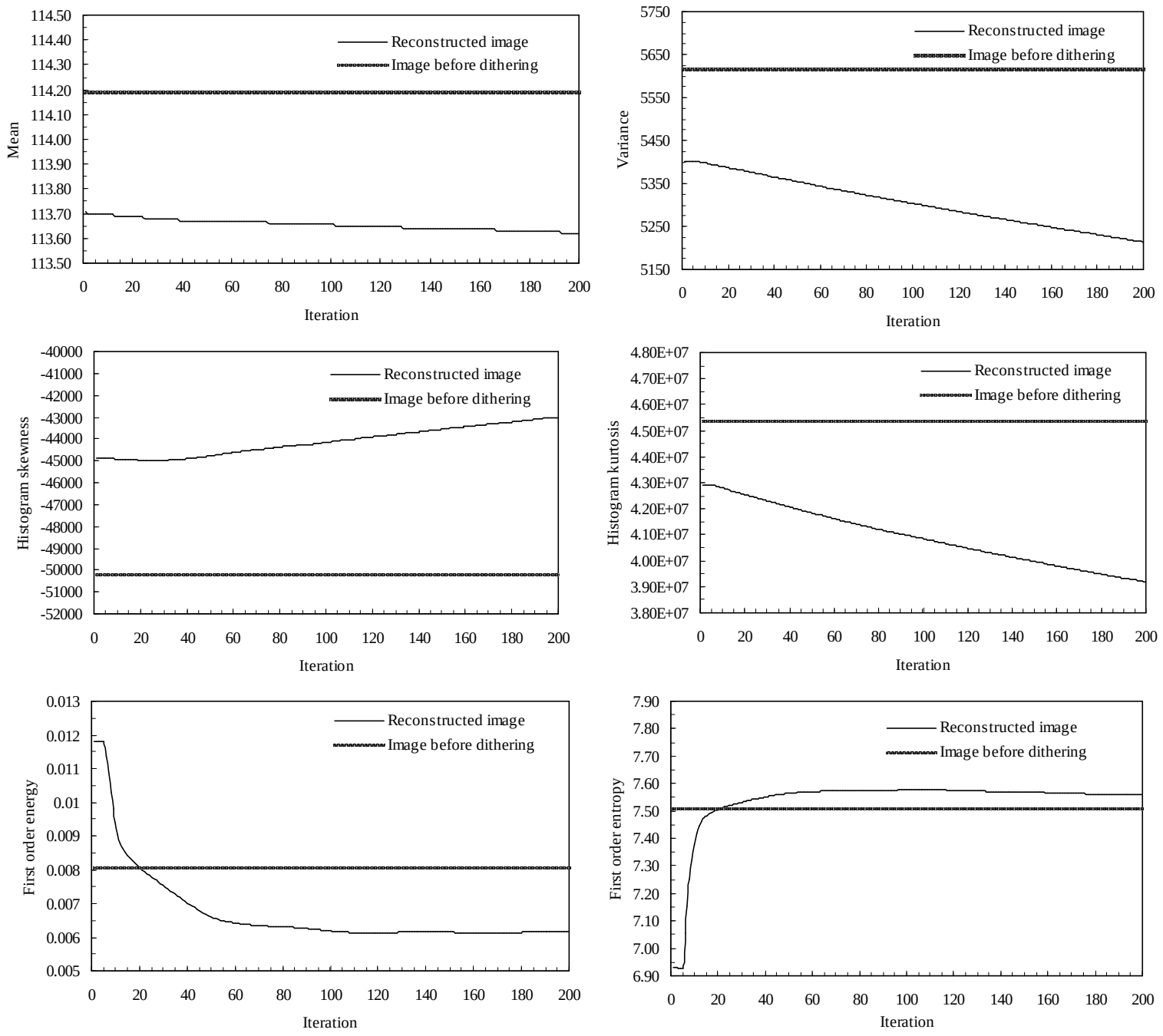

Fig. 3. First-order properties of the reconstructed image at each iteration against those of original pepper image

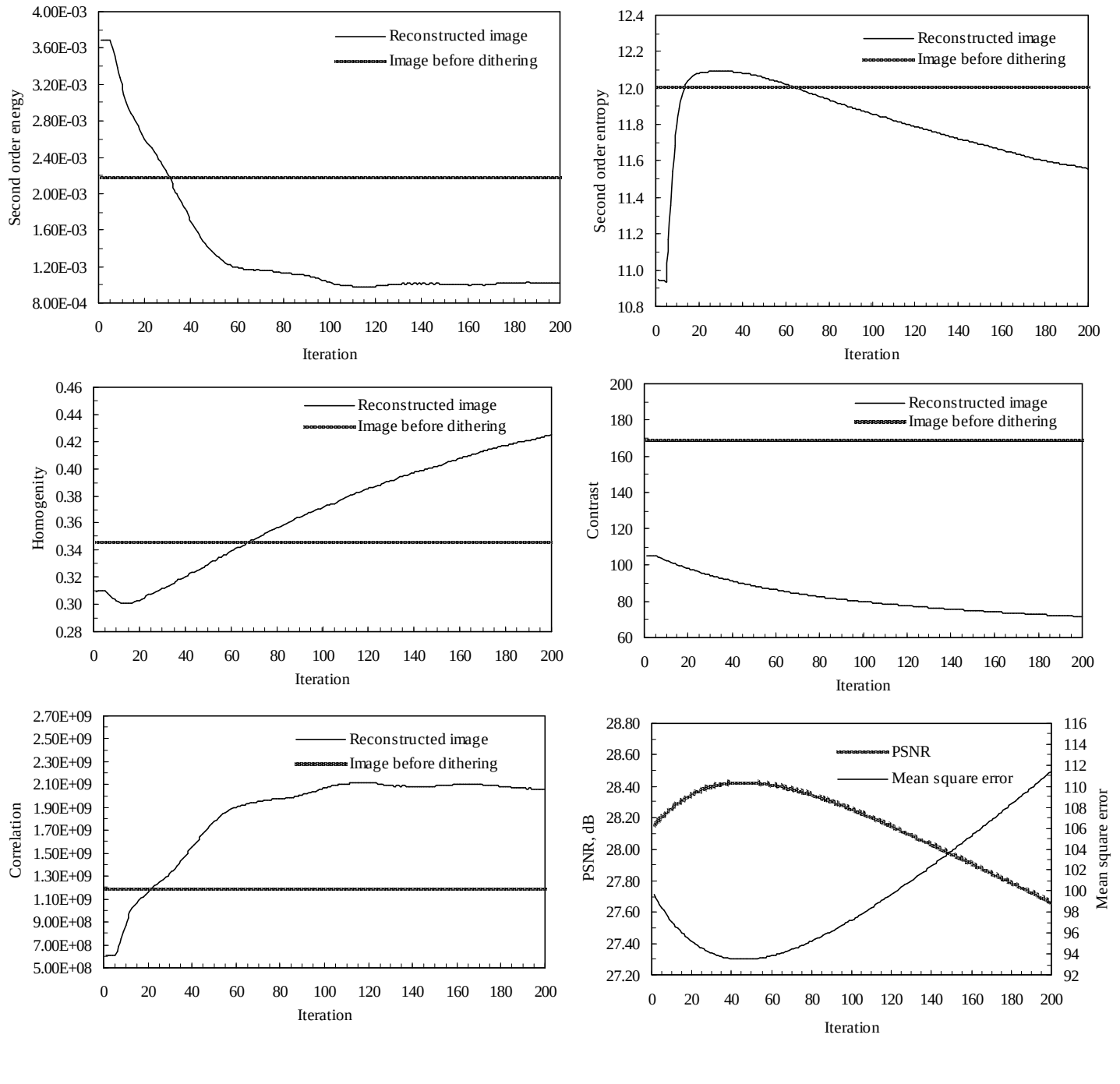

Fig. 4. Second-order properties of the reconstructed image at each iteration against those of original pepper image and MSE / PSNR

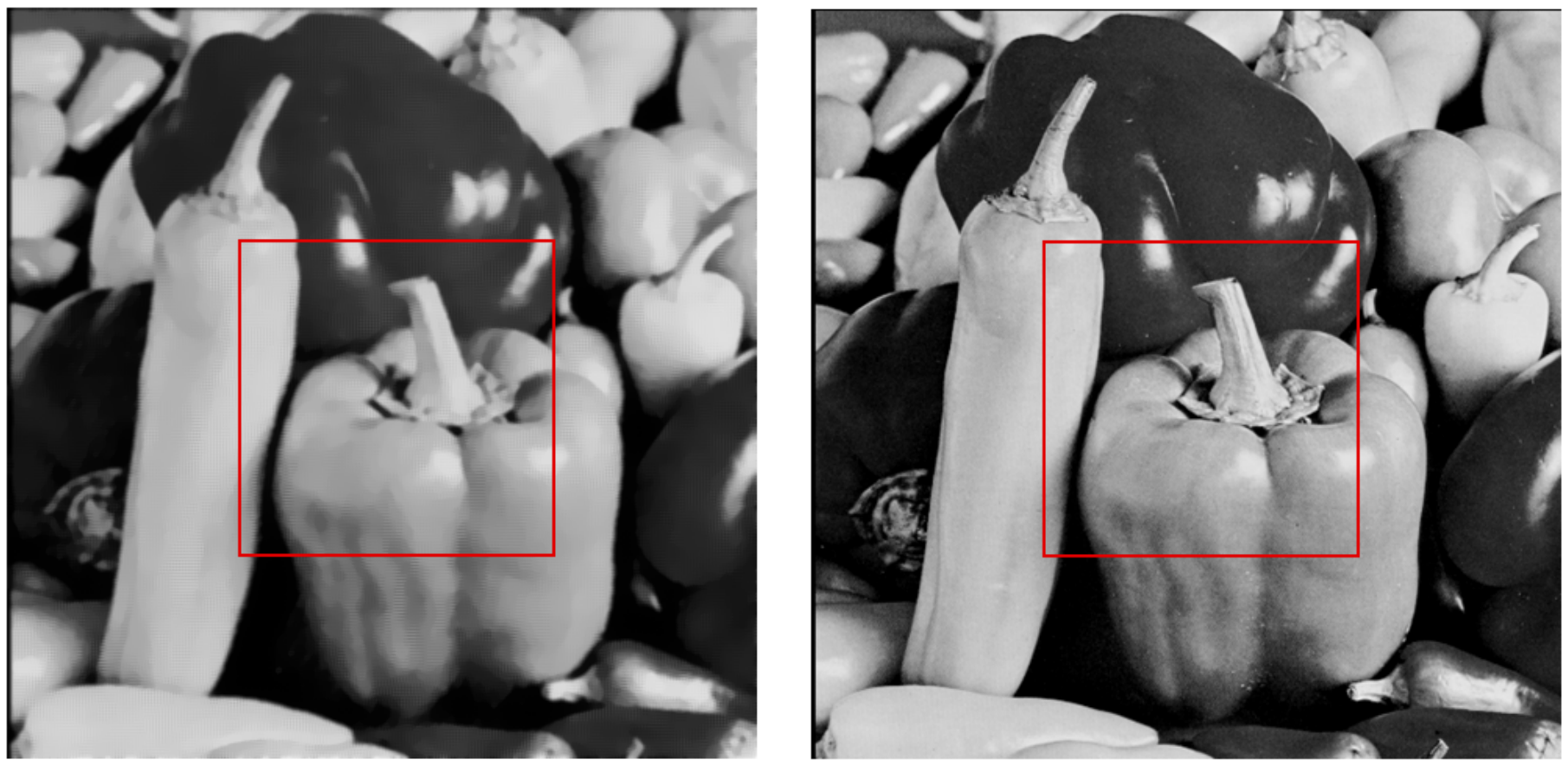

(a) (b)

Fig. 5. (a) Reconstructed pepper image (b) Original pepper image before dithering

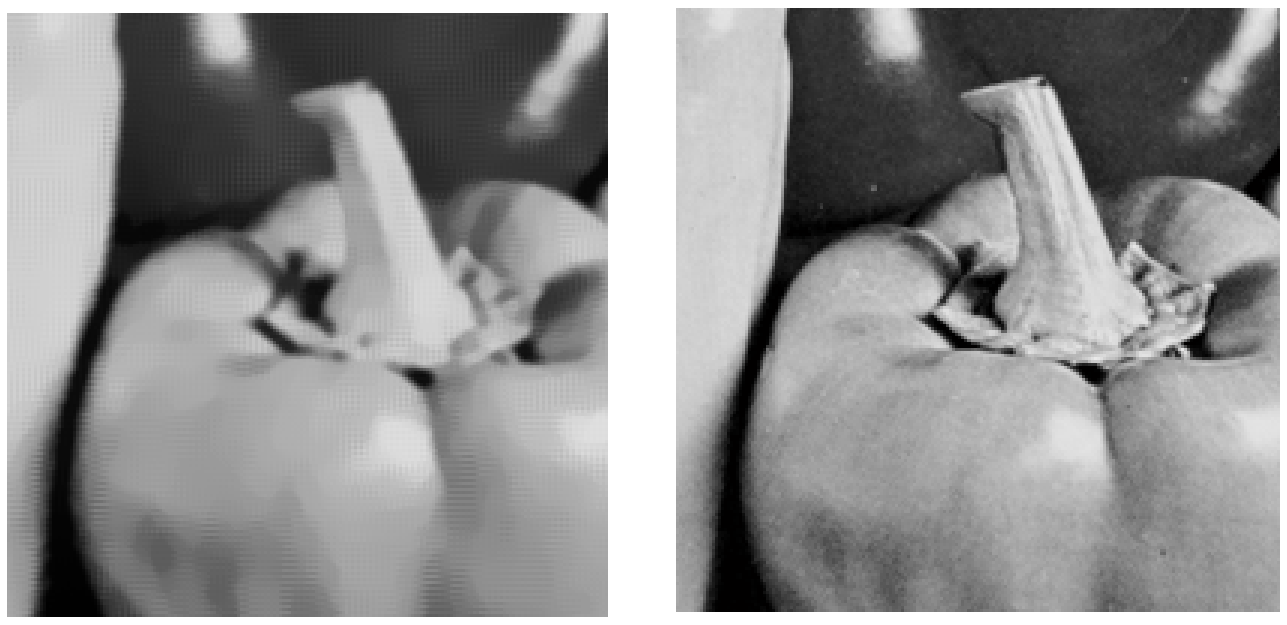

(a) (b)

Fig. 6. Detail of the window (a) Reconstructed image (b) Original pepper image before dithering

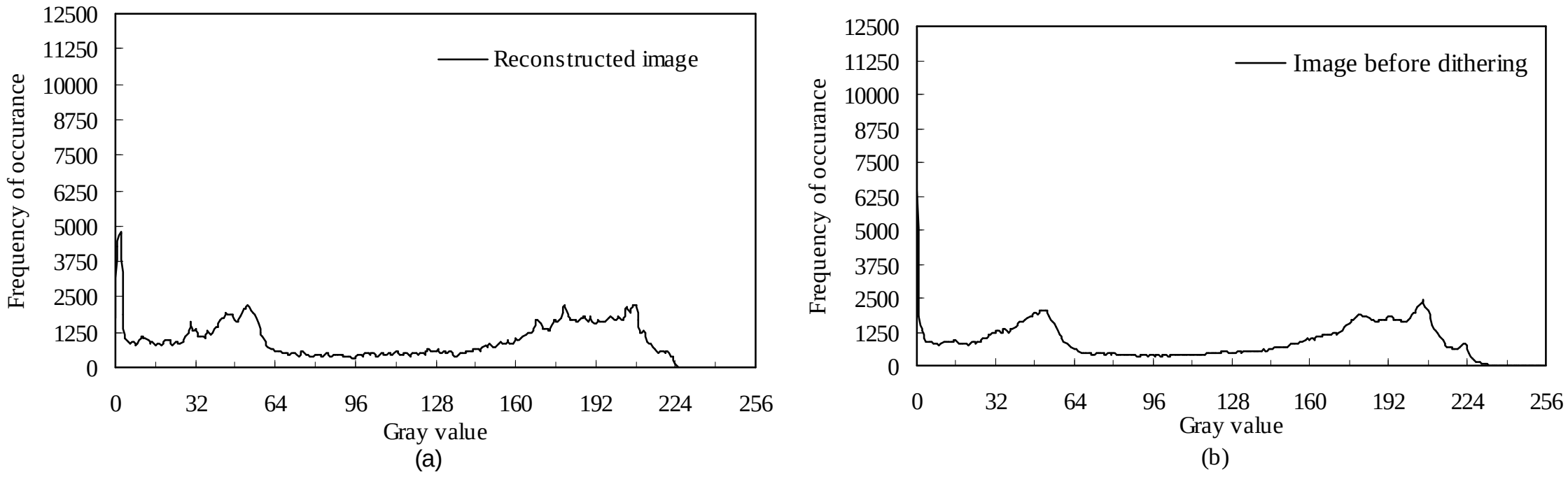

Fig. 7. Histograms (a) Reconstructed pepper image (b) Original pepper image before dithering

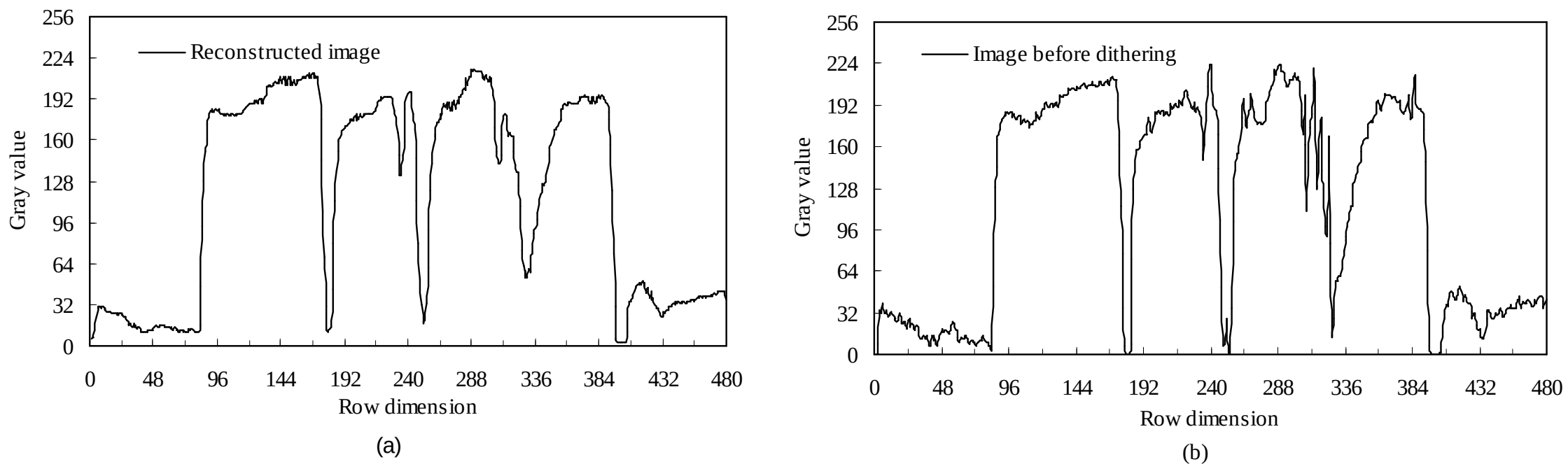

Fig. 8. Typical gray profiles (a) Reconstructed pepper image (b) Original pepper image before dithering

For the baboon image in dithered form as shown in Fig. 9, the features calculated are shown in Fig. 10 and Fig. 11. There is no minimum MSE at all. The MSE for this image keeps increasing with iteration. Moreover, at no iteration step, properties of the reconstructed image are close to those of the original image before dithering. Beyond an iteration step of 50, there is no much change in visual appearance and it is very difficult to define the quality of reconstruction. However, blurring effect slowly increases. Reconstructed image at a diffusion step of 100 appears somewhat better and is shown against the original image in Fig. 12. Detail of the small window of reconstructed image with reference to that of the original image shows the effect of smoothing as shown in Fig. 13. Comparison of histograms of reconstructed image and original image shows that a large number of gray values are reproduced with reference to those of original image as in Fig. 14. Typical gray profiles (Fig. 15) show that the gray level variation is also similar for both reconstructed image and original image.

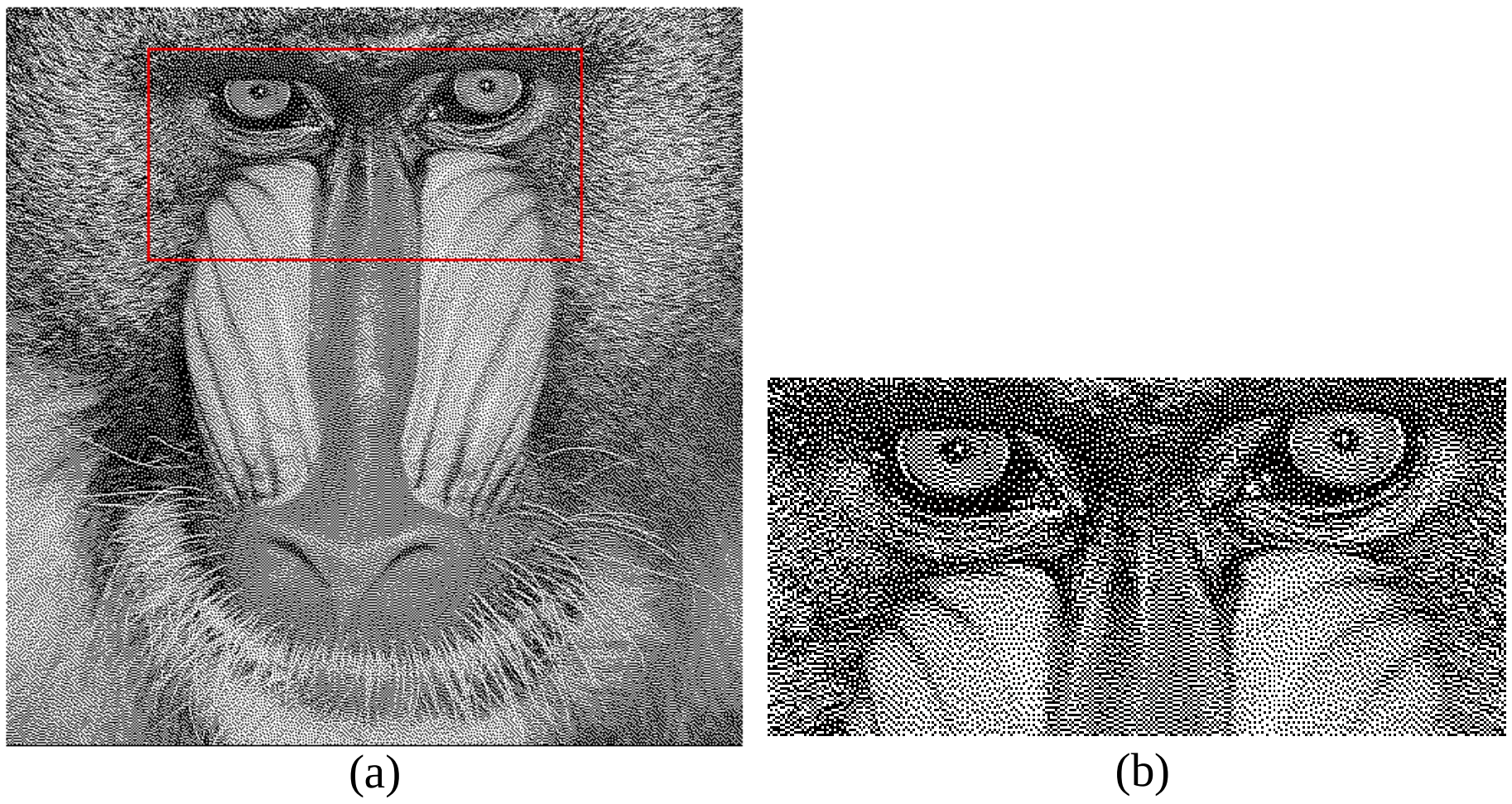

Fig. 9. Baboon image in dithered form (a) Full view (b) Detail of the window

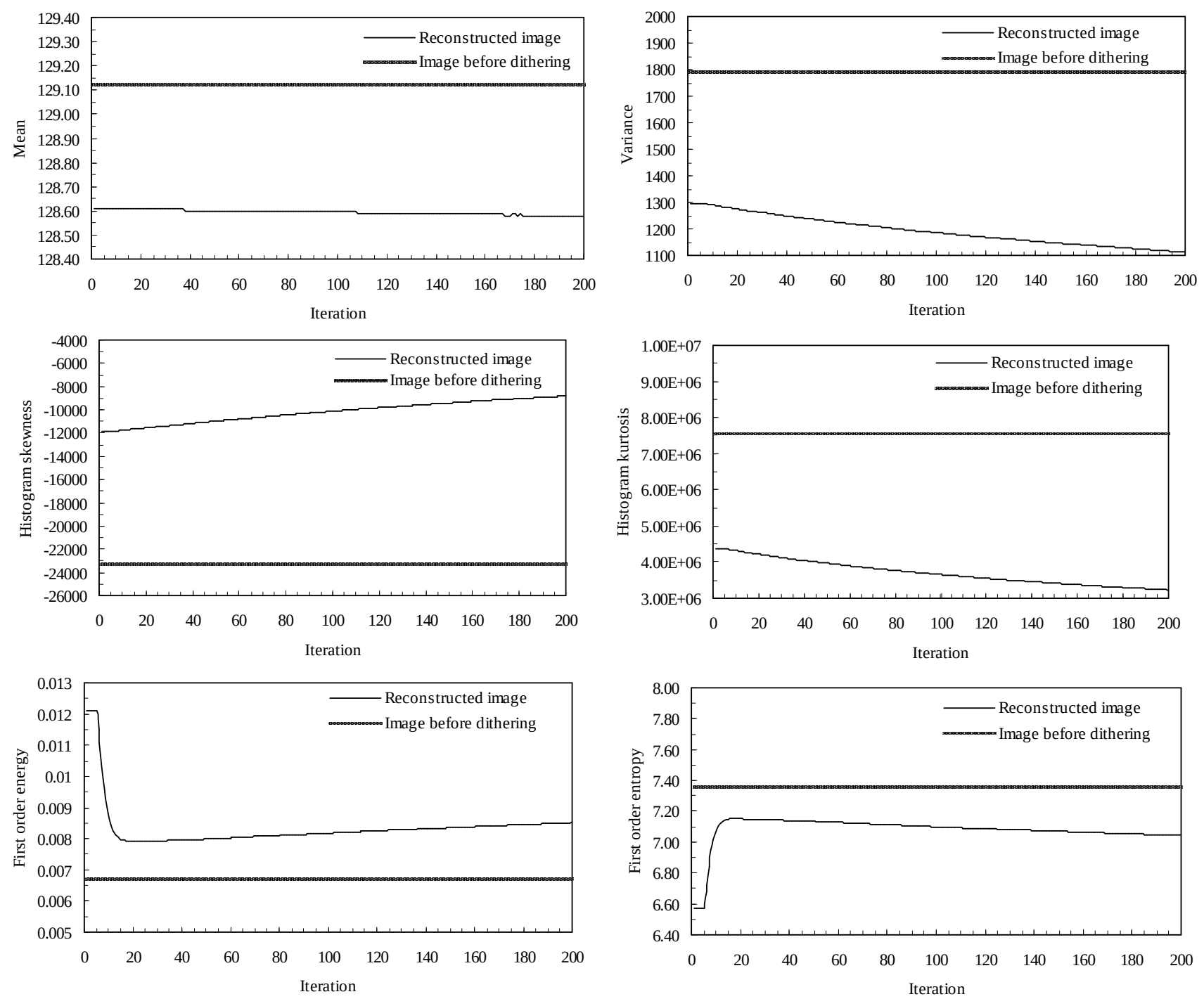

Fig. 10. First-order properties of the reconstructed image at each iteration against those of original baboon image

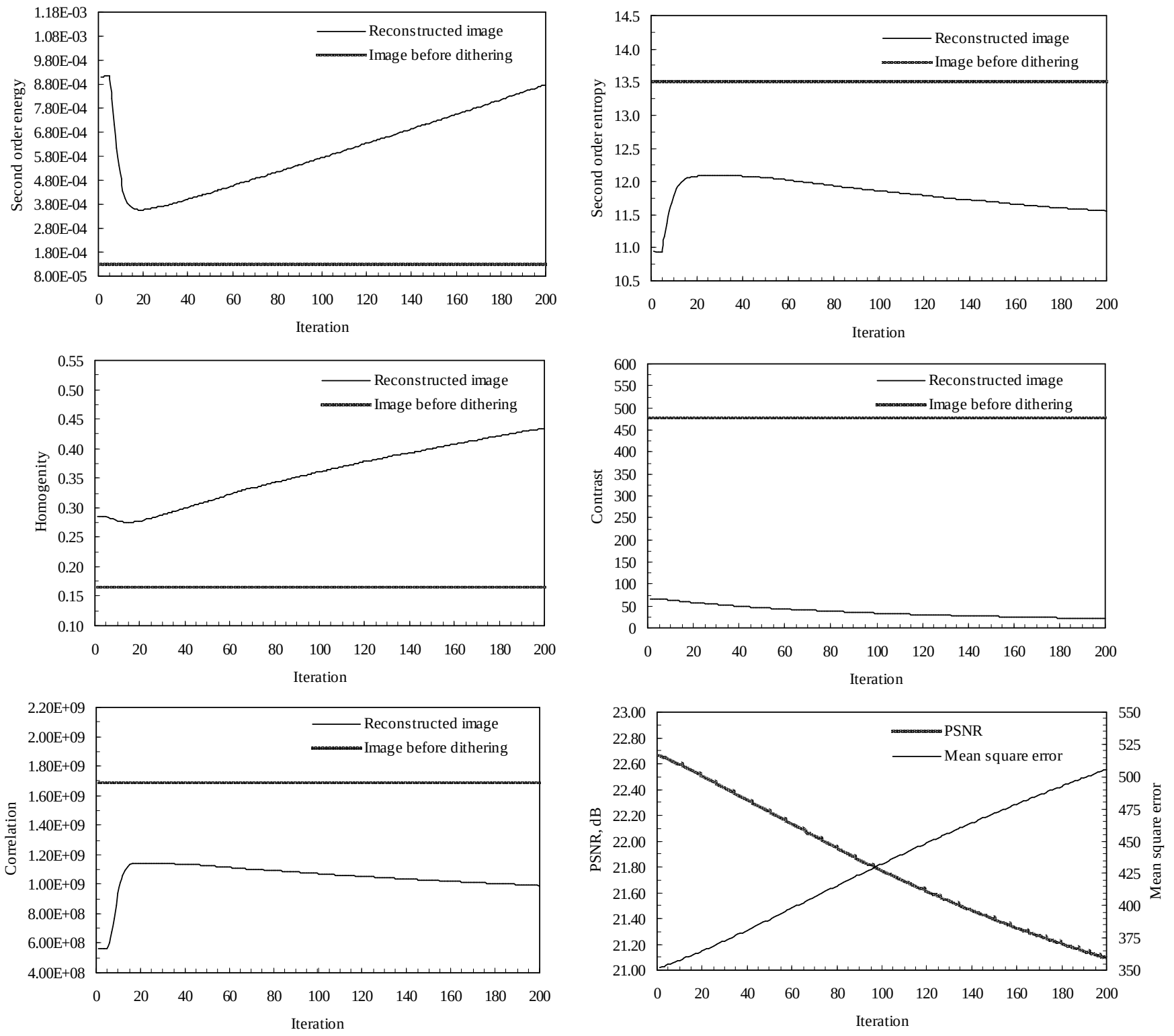

Fig. 11. Second-order properties of the reconstructed image at each iteration against those of original baboon image and MSE / PSNR

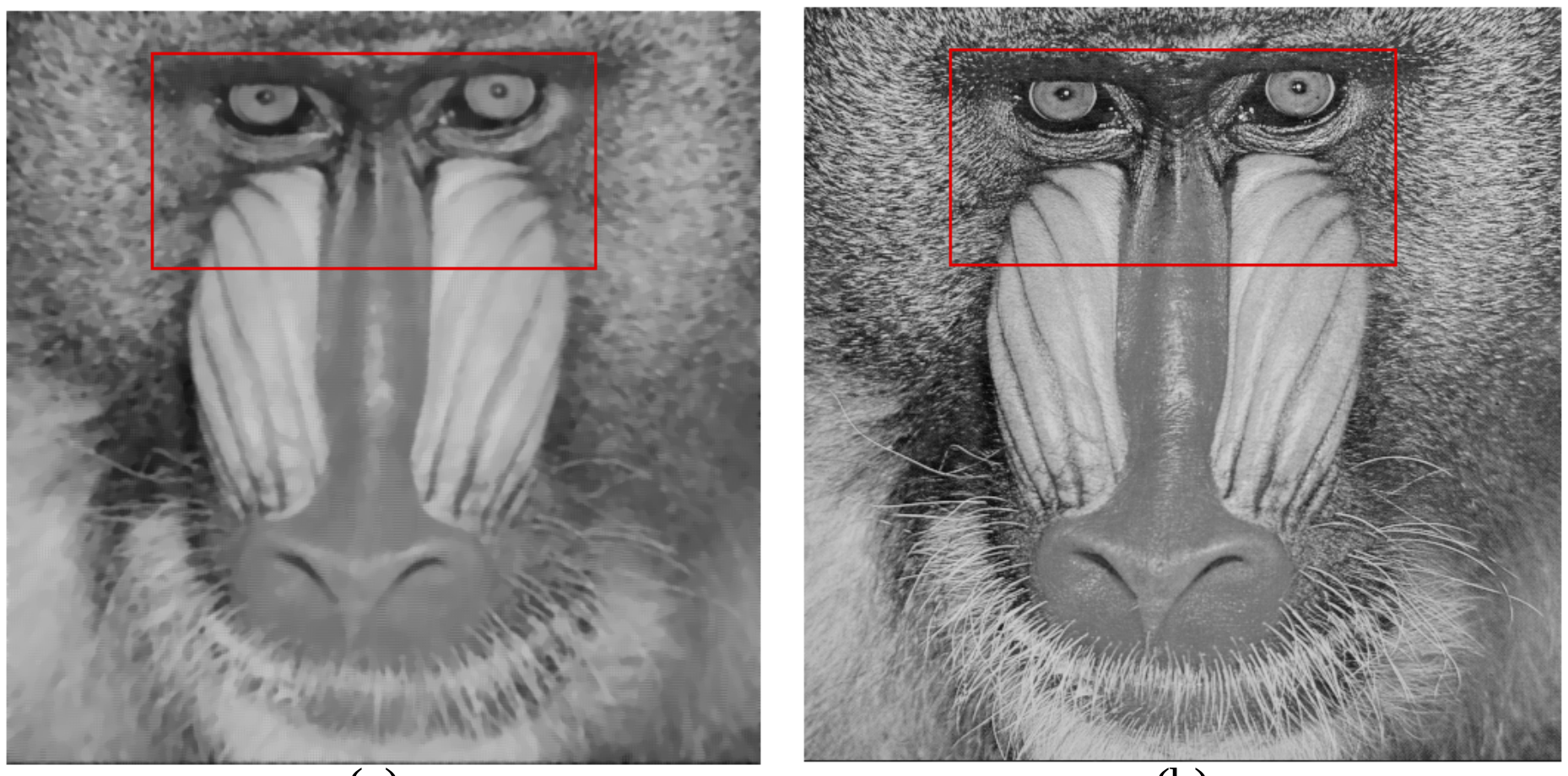

(a) (b)

Fig. 12. (a) Reconstructed baboon image (b) Original baboon image before dithering

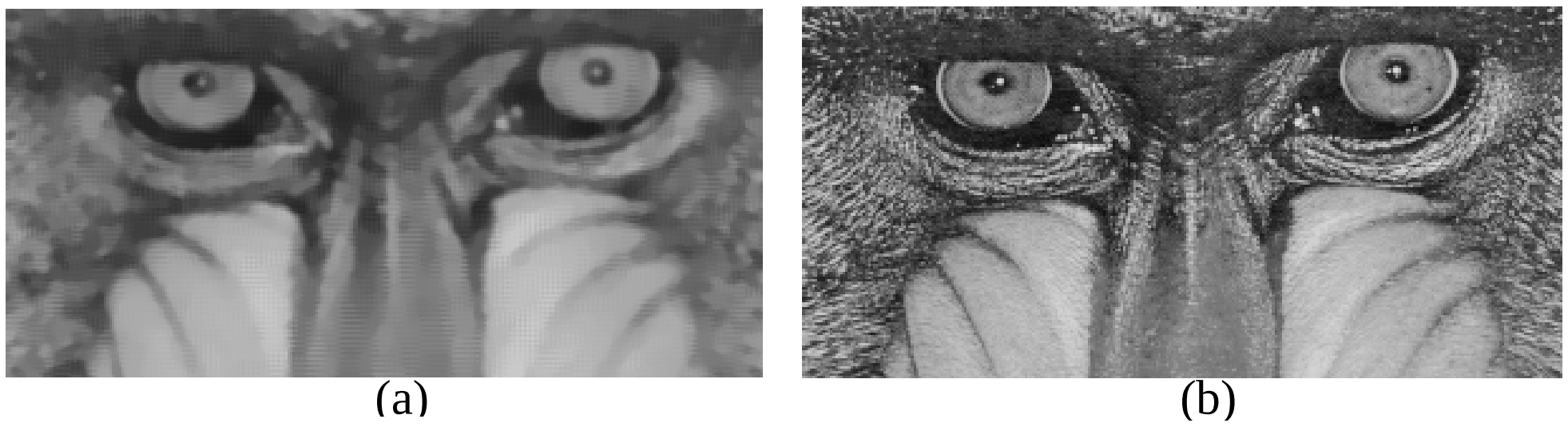

Fig. 13. Detail of the window (a) Reconstructed baboon image (b) Original baboon image before dithering

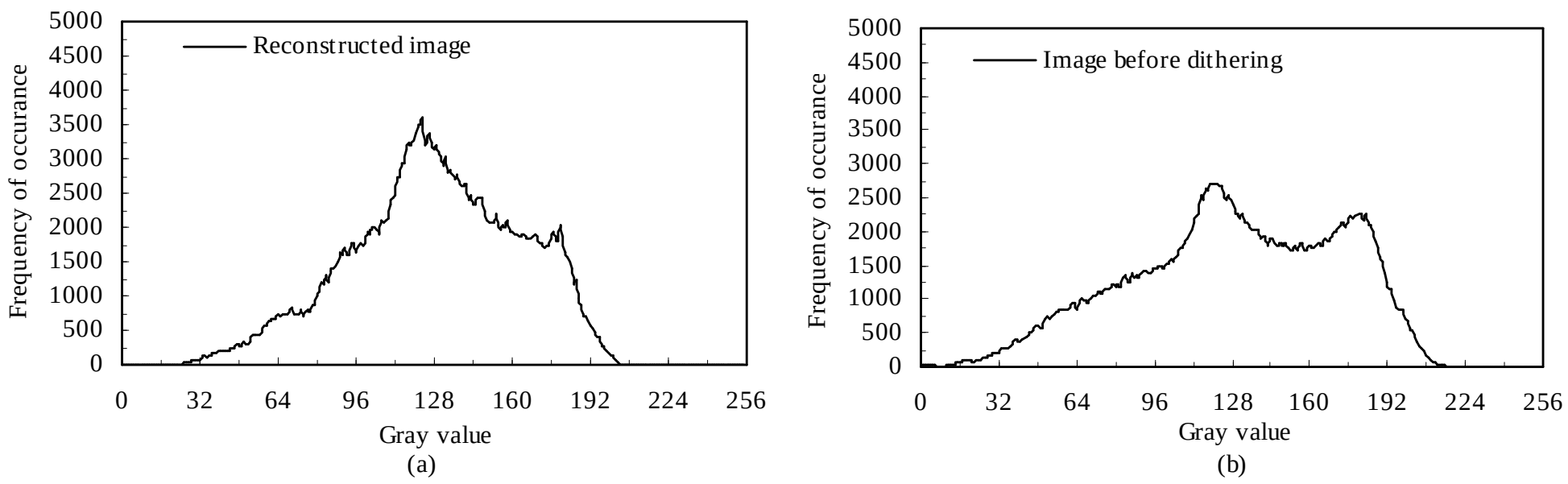

Fig. 14. Histograms (a) Reconstructed baboon image (b) Original baboon image before dithering

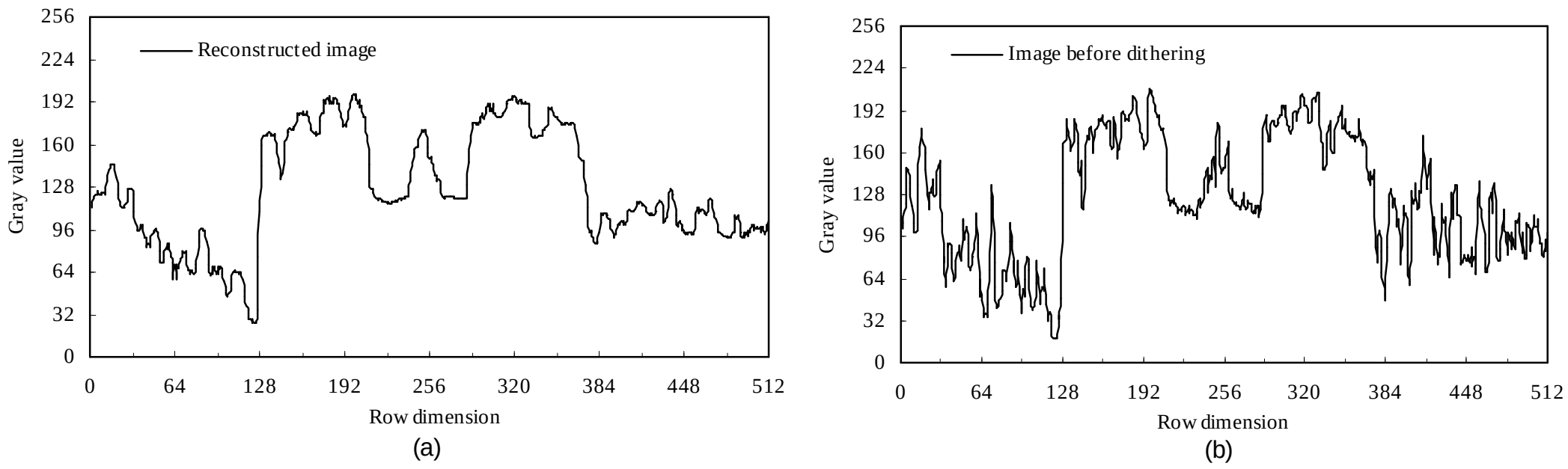

Fig. 15. Typical gray profile (a) Reconstructed baboon image (b) Original baboon image before dithering

## 5. SUMMARY AND CONCLUSIONS

It is shown in the present work that reconstruction of gray image from binary dithered image can be done using a combination of linear filtering and non-linear diffusion that brings the advantage of adaptive smoothing and edge enhancement. Histogram of the reconstructed image shows that a large number of gray values are reproduced in comparison with those of the original image before dithering. From an aesthetic point of view, the undithered images have similar appearance with reference to the original images but a little blurred.